%% file: main.tex
\documentclass[10pt]{article} 

\newif\ifsubmission
\usepackage[preprint]{tmlr}
\submissionfalse

\input{math_commands.tex}

\usepackage{hyperref}
\usepackage{url}
\usepackage{graphicx}
\usepackage{booktabs}
\usepackage{float}

\title{Fragile Knowledge, Robust Instruction-Following: The Width Pruning Dichotomy in Llama-3.2}

\author{\name Pere Martra \email peremartra@uadla.com \\
       \addr Independent Researcher}

\def\month{MM}  
\def\year{YYYY} 
\def\openreview{\url{https://openreview.net/forum?id=XXXX}} 

\begin{document}

\maketitle

\begin{abstract}
Structured width pruning of GLU-MLP layers in Llama-3.2 models, guided by the Peak-to-Peak Magnitude (PPM) criterion, reveals a systematic dichotomy in how reducing the expansion ratio affects different model capabilities. While performance on tasks relying on parametric knowledge (e.g., MMLU, GSM8K) and perplexity metrics degrades predictably with decreasing expansion ratios, instruction-following capabilities improve at the 2.4× equilibrium ratio (IFEval: $+4.8$ points / $+46\%$ in Llama-3.2-1B and $+3.7$ points / $+39\%$ in Llama-3.2-3B), and multi-step reasoning remains robust (MUSR). This pattern, observed consistently across both evaluated model sizes, challenges the prevailing assumption in compression research that pruning induces uniform degradation. To investigate this, we evaluated seven expansion ratio configurations using comprehensive benchmark suites that assess factual knowledge, mathematical reasoning, language comprehension, instruction-following, and truthfulness. Our analysis identifies the expansion ratio as a critical architectural parameter that selectively reshapes the model's task performance profile, rather than merely serving as a compression metric.  
\end{abstract}

\section{Introduction}
This study investigates the impact of structured width pruning on GLU-MLP layers \cite{guo_dependency-aware_2024}, using Llama-3.2 models as case studies. We analyze how the expansion ratio---a key architectural parameter---serves not only as a compression mechanism but also as an intervention that selectively modulates cognitive capabilities \cite{sharma_truth_2023}.

Building on a prior empirical observation \cite{martra_exploring_2024} that identified a $140\%$ expansion ratio as a performance equilibrium point, this work provides the first systematic characterization of this selective-degradation phenomenon. We document two core findings: (1) what we term a ``Capability Dichotomy''---observed under PPM selection in the base Llama-3.2-1B and 3B models---where pruning degrades parametric knowledge while preserving or improving instruction-following, and (2) an inverse correlation ($r = -0.864$) between MMLU and TruthfulQA-MC2, which we interpret as a scoring effect rather than a gain in truthfulness.

Section 1.1 motivates the need for efficient models, Section 1.2 contextualizes width pruning within LLM optimization techniques, Section 1.3 details our contributions, and Section 1.4 previews key results.

\subsection{Motivation: The Need for Efficient Models}
Large language models have demonstrated unprecedented capabilities across a wide range of tasks \cite{zhao_survey_2025}, but their increasing size incurs significant computational and energy costs during both training and inference. Models with tens or hundreds of billions of parameters require specialized infrastructure and consume substantial resources, which limits their accessibility and sustainability \cite{muralidharan_compact_2024}.

Reducing the computational footprint of LLMs is no longer merely an academic goal but a practical necessity to democratize access, enable deployment on resource-constrained devices (e.g., edge devices), and ensure long-term economic and environmental viability. This urgent need for efficiency has spurred research into techniques for improving the efficiency of large language model-based solutions \cite{sun_simple_2024}.

Among these techniques, structured pruning---the systematic removal of neural components---has emerged as a particularly promising approach, traditionally viewed as a compression method \cite{muralidharan_compact_2024, xia_sheared_2024}. However, applying width pruning to GLU layers introduces a fundamental architectural question: Does capacity reduction uniformly degrade all cognitive functions, or can it induce selective changes?. In this study, we use pruning not only as a compression tool but also as an analytical lens to explore this question, systematically examining how variations in the expansion ratio reshape the model's capability profile.

\subsection{Width Pruning in the Context of LLM Optimization}
Structured pruning is one of several strategies employed by the research community for model optimization. It complements two other primary approaches: quantization, which reduces the numerical precision of model weights (e.g., from 16-bit to 4-bit representation), and knowledge distillation, which trains a smaller "student" model to replicate the behavior of a larger "teacher" model \cite{muralidharan_compact_2024}.

Unlike quantization and distillation---methods that operate at the representation and training levels, respectively---structured pruning directly modifies the model's architecture by removing entire components. This approach permanently reduces the parameter count \cite{sun_simple_2024, xia_sheared_2024}, enabling both the elimination of inherent structural redundancies in pre-trained models and the adaptation of general-purpose models to task-specific deployments. By prioritizing the most relevant capabilities for a given use case, pruning can optimize models for efficiency without sacrificing critical functionality \cite{reda_how_2025}.

Structured pruning techniques are broadly categorized into depth pruning (removing entire layers) and width pruning (reducing layer dimensions) \cite{kim_shortened_2024, muralidharan_compact_2024}. While depth pruning represents an aggressive and coarse-grained intervention, width pruning offers finer-grained control, enabling more precise modifications to the model's behavior \cite{sharma_truth_2023, wei_assessing_2024}. This study focuses on the systematic application of width pruning to adjust the intermediate dimension of MLP layers, thereby altering the expansion ratio---a fundamental yet understudied architectural parameter whose impact on model capabilities remains poorly understood.

\subsection{Contributions}
This work presents a systematic analysis of the impact of width pruning on GLU architectures, using Llama-3.2-1B and 3B models as case studies. Specifically, we examine how variations in the expansion ratio differentially affect cognitive capabilities in language models. Our main contributions are as follows:

\begin{itemize}
    \item \textbf{Dichotomy of capabilities under the PPM criterion:} We show 
that width pruning guided by the Peak-to-Peak Magnitude (PPM) criterion 
affects different task types in distinct ways. While capabilities reliant 
on parametrized knowledge---such as performance on MMLU, GSM8K, and 
perplexity---degrade predictably with reductions in the expansion ratio 
\cite{muralidharan_compact_2024}, instruction-following metrics (IFEval) 
improve at the $2.4\times$ ratio ($+4.8$ points / $+46\%$ in Llama-1B 
and $+3.7$ points / $+39\%$ in Llama-3B, from baselines of $0.104$ and 
$0.094$ respectively; see Appendix~\ref{sec:completebenchmarks}). This 
reveals a systematic trade-off between factual memory and behavioral 
adherence. Notably, this pattern is consistent across both 1B and 3B 
models and is specific to the PPM criterion. Alternative weight-only 
criteria---the standard L2 norm, Variance of Weights (VOW), and Product 
of L1 Norms (PON)---do not exhibit this behavior and instead result in 
catastrophic performance collapse (see Appendix~\ref{sec:ppmjustification}).
    \item \textbf{Efficiency trade-offs by inference mode:} We quantify the efficiency trade-offs of pruning, which consistently reduces energy consumption by up to $23\%$ (J/token) but introduces end-to-end latency penalties that depend on the operational context \cite{gholami_can_2023, muralidharan_compact_2024}. In single-request configurations (B1), we observe latency increases of up to $+18\%$, whereas batch processing (B8) shows resilient throughput. These findings indicate that pruned configurations are better optimized for high-concurrency workloads than for interactive applications.
\end{itemize}

Additionally, we complement this analysis with a detailed characterization of the carbon footprint (see Appendix C) and an evaluation of instruct-tuned models (see Appendix B). In these evaluations, an expansion ratio of $2.4\times$ emerges as an equilibrium point, balancing competitive capabilities across both model sizes.

\subsection{Results Preview}
Our experiments show that PPM-guided width pruning in GLU-MLP layers is more than a uniform compression technique; it is an intervention that selectively reshapes the model's cognitive capabilities \cite{hou_instruction-following_2025, sharma_truth_2023, wei_assessing_2024}. The results reveal a systematic dichotomy: while performance on tasks dependent on parameterized factual knowledge---such as MMLU and GSM8K---degrades predictably as the expansion ratio decreases, instruction-following metrics improve substantially. Specifically, at the shared $2.4\times$ equilibrium ratio, IFEval increases by $+4.8$ points ($0.104 \rightarrow 0.152$, $+46\%$) in Llama-3.2-1B and $+3.7$ points ($0.094 \rightarrow 0.131$, $+39\%$) in Llama-3.2-3B. Larger peak gains occur at more aggressive ratios, but at a different expansion ratio in each model (see Section~\ref{sec:dicothomy} and Appendix~\ref{sec:completebenchmarks}).

Additionally, our analysis uncovers a critical efficiency trade-off that depends on the inference mode \cite{gholami_can_2023, muralidharan_compact_2024}. Pruning consistently reduces energy consumption (J/token) but incurs significant end-to-end latency penalties in single-request configurations (B1). However, these latency costs are substantially mitigated in batch processing scenarios (B8), suggesting that the architectural bottleneck affecting single-stream generation does not constrain parallel processing capacity.

Finally, our study identifies an expansion ratio of $2.4\times$ as an optimal balance point for the evaluated models, effectively reconciling the competing objectives of capability preservation and efficiency.

\section{Background \& Related Work}
This section establishes the technical foundations for our study. Section 2.1 describes the Gated Linear Unit (GLU) architecture, a core component of our pruning methodology. Section 2.2 positions our work within the broader context of structured pruning research.

\subsection{GLU Architecture}
Gated Linear Unit (GLU) layers represent an evolution of the standard Feed-Forward Network (FFN) layers in transformer models. While a vanilla FFN applies a simple transformation of the form $h=\sigma(xW_1)W_2$, GLU introduces a gating mechanism that modulates information flow through element-wise multiplication \cite{shazeer_glu_2020}.

\begin{figure}[ht]
    \centering
    \includegraphics[width=0.6\textwidth, 
                       clip, 
                       trim={0cm 1.3cm 0.5cm 0.3cm}]{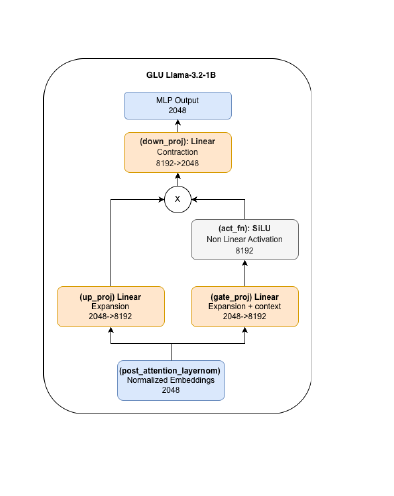}]
    
    \caption{GLU (Gated Linear Unit) architecture within the MLP block of Llama-3.2-1B. The diagram shows the input flow from \texttt{post\_attention\_layernom}, the parallel \texttt{up\_proj} and \texttt{gate\_proj} projections, the SiLU activation function (\texttt{act\_fn}), the element-wise multiplication (gating), and the final \texttt{down\_proj} contraction.}
    
    \label{fig:glu-architecture}
\end{figure}

As illustrated in Figure 1, the transformation in a GLU layer is defined as:
$$h = (xW_{\text{up}} \odot \text{SiLU}(xW_{\text{gate}}))W_{\text{down}} \quad (1)$$
where:
\begin{itemize}
    \item $x \in \mathbb{R}^{d_{\text{model}}}$ is the post-attention input
    \item $W_{\text{up}}, W_{\text{gate}} \in \mathbb{R}^{d_{\text{model}} \times d_{\text{ff}}}$ are the parallel expansion projections
    \item $W_{\text{down}} \in \mathbb{R}^{d_{\text{ff}} \times d_{\text{model}}}$ is the contraction projection
    \item $\odot$ denotes the element-wise (Hadamard) product
    \item $\text{SiLU}(x) = x \cdot \sigma(x)$ is the Sigmoid Linear Unit activation function
\end{itemize}

The expansion ratio is defined as $r = d_{\text{ff}} / d_{\text{model}}$ and determines the MLP layer capacity. For Llama-3.2-1B, $d_{\text{model}} = 2048$ and $d_{\text{ff}} = 8192$, resulting in $r = 4.0\times$. For Llama-3.2-3B, $d_{\text{model}} = 3072$ and $d_{\text{ff}} = 8192$, giving $r \approx 2.67\times$.

A fundamental characteristic of GLU for structured pruning is that $W_{\text{up}}$ and $W_{\text{gate}}$ must maintain identical dimensions due to the gating mechanism \cite{guo_dependency-aware_2024}. This means that any reduction in $d_{\text{ff}}$ must be applied in a paired manner to both projections simultaneously, removing the same neurons in both layers. This paired pruning constraint is essential to maintain the architectural coherence of the pruned model \cite{guo_dependency-aware_2024}.

\subsection{Related Work}
Structured pruning has been explored through various strategies to reduce the size of LLMs while preserving their performance. These techniques can be broadly classified into two categories: width pruning, which reduces the dimensionality of layers, and depth pruning, which removes entire layers.

\textbf{Width Pruning in LLMs.} Several recent studies address structured pruning at the neuron or channel level. \textbf{SliceGPT} \cite{ashkboos_slicegpt_2024} employs principal component analysis on activation matrices to identify neurons for removal through channel-wise pruning, applying transformations at the full block level. However, this activation-based approach exhibits high sensitivity to the calibration data used. \textbf{AMP} \cite{mugnaini_efficient_2025} introduces a method that prunes attention heads and MLP neurons simultaneously by projecting input data onto weights, assessing structural importance through activation magnitudes. Although flexible, AMP does not systematically explore the effects of varying expansion ratios in MLP layers. As noted in comprehensive surveys (e.g., Zhao et al., 2025) \cite{zhao_survey_2025}, width pruning remains relatively underexplored compared to techniques like quantization or knowledge distillation, particularly in the context of systematic architectural analysis.

\textbf{Depth Pruning.} Unlike width pruning, the literature on depth pruning (removal of entire transformer blocks) presents conflicting results. Some works, such as Kim et al. \cite{kim_shortened_2024}, advocate for this technique, arguing that it can achieve "comparable or superior" performance to width pruning. However, more recent research, such as Wang et al. \cite{wang_when_2025}, demonstrates that removing even a few layers severely degrades test-time scaling, causing a "catastrophic collapse" in long-chain reasoning benchmarks---a structural damage that proves irrecoverable through fine-tuning. Our width pruning approach offers more granular control compared to this drastic measure, enabling more gradual and selective degradation.

\textbf{GLU-Specific Pruning.} Prior work that systematically analyzes the impact of expansion ratios in GLU architectures is limited. A previous exploratory study \cite{martra_exploring_2024} identified a $140\%$ expansion ratio as an empirical equilibrium point for performance in Llama-3.2 models. However, that work did not provide a systematic characterization of this selective-degradation phenomenon or its impact on different cognitive capabilities. To our knowledge, no study has framed the expansion ratio as an intervention to selectively modulate cognitive functions. This gap---moving from empirical observation to systematic characterization---motivates our approach: we show that width pruning in GLU-MLP layers does not simply constitute uniform compression but rather an intervention that selectively modifies different cognitive capabilities, with the expansion ratio emerging as a critical metric for determining pruning effectiveness.

\section{Methodology}
\label{sec:methodology}
This section describes the methodological approach used to systematically prune Llama-3.2 models and evaluate the impact of varying GLU expansion ratios on model performance and capabilities.

\subsection{Width Pruning in GLU Architectures}
Width pruning reduces the intermediate dimension of MLP layers in transformers, directly modifying the architectural parameter known as the expansion ratio. In architectures based on Gated Linear Units (GLU), each MLP layer contains three linear projections: $\text{gate\_proj}$, $\text{up\_proj}$, and $\text{down\_proj}$ \cite{shazeer_glu_2020}.

The first two process the input in parallel (with an output dimension of $\text{intermediate\_dim}$), where $\text{gate\_proj}$ applies a sigmoid activation that modulates the output of $\text{up\_proj}$ before projecting back to the hidden dimension through $\text{down\_proj}$. The expansion ratio is defined as $\text{intermediate\_dim} / \text{hidden\_dim}$ and determines the expansion capacity of these layers.

Width pruning in GLU requires \textbf{paired pruning}: neurons removed from $\text{gate\_proj}$ must exactly correspond to those removed from $\text{up\_proj}$ to maintain the coherence of the gating mechanism \cite{guo_dependency-aware_2024}.

\textbf{Neuron Selection: PPM} To determine which neurons to prune, we conducted a preliminary evaluation comparing four weight-only importance criteria: Peak-to-Peak Magnitude (PPM), the standard L2 norm, Variance of Weights (VOW), and Product of L1 Norms (PON) (see Appendix D). The L2, VOW, and PON criteria resulted in catastrophic performance degradation immediately after pruning.

The PPM method calculates the importance of each neuron as the peak-to-peak magnitude of its incoming weights. This calculation accounts for the parallel operation of the $\text{up\_proj}$ and $\text{gate\_proj}$ layers.

The formula used to calculate the importance score ($\text{importance\_scores}$) of each candidate neuron, combining the weight values from the two paired layers, is as follows:

\textbf{Step 1: Calculation of the Peak-to-Peak Magnitude for each expansion layer}
$$\text{gate\_max\_abs} = \max(W_{\text{gate}}, \text{axis}=1) + |\min(W_{\text{gate}}, \text{axis}=1)|$$
$$\text{up\_max\_abs} = \max(W_{\text{up}}, \text{axis}=1) + |\min(W_{\text{up}}, \text{axis}=1)|$$

\textbf{Step 2: Calculation of the Combined Importance Score}
$$\text{importance\_scores} = \text{gate\_max\_abs} + \text{up\_max\_abs}$$

The neurons with the lowest importance scores, as determined by this formula, are selected for removal from the model.

The selection of PPM as the importance method is justified by preliminary 
results documented in Appendix D. The three alternative criteria---the 
standard L2 norm, Variance of Weights (VOW), and Product of L1 Norms 
(PON)---resulted in catastrophic performance collapse at $10\%$ pruning 
in Llama-3.2-1B: all three drove LAMBADA perplexity into the thousands 
of percent above baseline, with L2 and PON additionally causing WikiText 
perplexity to diverge. By contrast, PPM maintained moderate increases 
($+54\%$ in WikiText, $+230\%$ in LAMBADA), confirming its suitability 
for gradual pruning in GLU architectures (see Appendix~\ref{sec:ppmjustification}).

The pruning implementation was carried out using the \textbf{optipfair} library (v0.2.0) \cite{martra_optipfair_2024}, a specialized tool for structured pruning in GLU architectures that automatically handles paired pruning and ensures dimensional consistency \cite{guo_dependency-aware_2024}. The configuration applied uniform pruning across all model layers while preserving the attention architecture. All operations were performed using $\text{bfloat}16$ precision.

\subsection{Experimental configuration}
We evaluated two models from the Llama-3.2 family \cite{meta_ai_llama-modelsmodelsllama3_2model_cardmd_2024} in their base version (not instruct):

\begin{itemize}
    \item Llama-3.2-1B: $\text{hidden\_size}=2048$, $\text{intermediate\_size}=8192$, baseline expansion ratio of $4.0\times$
    \item Llama-3.2-3B: $\text{hidden\_size}=3072$, $\text{intermediate\_size}=8192$, baseline expansion ratio of $2.67\times$
\end{itemize}

This selection enables us to analyze how pruning affects models with different baseline expansion ratios and parameter scales.

For each model, we evaluated seven pruning configurations, ranging from the unpruned baseline ($0\%$) to aggressive pruning ($60\%$). Since the models have different baseline expansion ratios, the same pruning percentage results in different final expansion ratios. Table 1 presents the complete mapping:

\begin{table}[H]
\caption{Pruning configurations and resulting expansion ratios for Llama-3.2-1B and 3B models.}
\label{tab:pruning-configs}
\centering
\begin{tabular}{lcccc}
\toprule
\textbf{Expansion} & \textbf{1B Pruning} & \textbf{3B Pruning} & \textbf{Inter. Dim} & \textbf{Inter. Dim} \\
\textbf{Ratio} & \textbf{(\%)} & \textbf{(\%)} & \textbf{(1B)} & \textbf{(3B)} \\
\midrule
4.0x   & 0\% (baseline) & -- & 8192 & -- \\
3.6x   & 10\% & -- & 7373 & -- \\
3.2x   & 20\% & -- & 6554 & -- \\
2.8x   & 30\% & -- & 5735 & -- \\
2.67x  & -- & 0\% (baseline) & -- & 8192 \\
2.4x   & 40\% & 10\% & 4916 & 7373 \\
2.13x  & -- & 20\% & -- & 6554 \\
2.0x   & 50\% & -- & 4096 & -- \\
1.87x  & -- & 30\% & -- & 5735 \\
1.6x   & 60\% & 40\% & 3277 & 4916 \\
1.33x  & -- & 50\% & -- & 4096 \\
1.07x  & -- & 60\% & -- & 3277 \\
\bottomrule
\end{tabular}
\end{table}

We used the expansion ratio as the primary independent variable to facilitate cross-model comparisons, as it directly represents the architectural capacity of MLP layers regardless of the model's base size \cite{shazeer_glu_2020}.

The pruned models were evaluated on a suite of 13 benchmarks covering diverse cognitive aspects of language models \cite{muralidharan_compact_2024}:

\begin{table}[H]
\caption{Benchmark suite employed for evaluation across cognitive categories.}
\label{tab:benchmark-suite}
\centering
\begin{small} 
\begin{tabular}{llcll}
\toprule
\textbf{Category} & \textbf{Benchmark} & \textbf{Shots} & \textbf{Metric} & \textbf{Description} \\
\midrule
Knowledge & MMLU & 5 & Accuracy & Multidisciplinary knowledge \\
          & ARC-Challenge & 0 & Accuracy & Scientific reasoning \\
\addlinespace 
Math      & GSM8K & 5 & Exact Match & Math problems with CoT \\
Reasoning & MUSR & 0 & Acc-Norm & Multi-step reasoning \\
\addlinespace
Language  & HellaSwag & 0 & Acc-Norm & Sentence completion \\
Understanding & WinoGrande & 0 & Accuracy & Ambiguity resolution \\
          & PIQA & 0 & Accuracy & Physical reasoning \\
          & BoolQ & 0 & Accuracy & Boolean questions \\
\addlinespace
Language  & WikiText & 0 & Perplexity & Continuous text perplexity \\
Modeling  & Lambada & 0 & Perplexity & Last word prediction \\
\addlinespace
Truthfulness & TruthfulQA-MC1 & 0 & Accuracy & Truthfulness (single correct) \\
             & TruthfulQA-MC2 & 0 & Accuracy & Truthfulness (multi-correct) \\
\addlinespace
Instruction & IFEval & 0 & Strict Acc & Instruction adherence \\
Following   & & & & \\
\bottomrule
\end{tabular}
\end{small}
\end{table}

To quantify efficiency trade-offs (Contribution 3), we evaluated inference in two operational scenarios:

\begin{itemize}
    \item \textbf{Single-Request ($\text{batch\_size}=1$):} Simulating interactive applications, we measured end-to-end latency (time to complete generation) and energy consumption (Joules/token).
    \item \textbf{Batch Processing ($\text{batch\_size}=8$):} Simulating high-concurrency workloads, we measured throughput and energy consumption (Joules/token).
\end{itemize}

These measurements were performed using \textbf{CodeCarbon} \cite{benoit_courty_mlco2codecarbon_2025} on a set of representative benchmarks:

\begin{itemize}
    \item HellaSwag (20 tokens, short generation)
    \item MMLU (50 tokens, knowledge responses)
    \item IFEval (150 tokens, instruction following)
\end{itemize}

These benchmarks were configured to ensure diversity in generation lengths and task types \cite{muralidharan_compact_2024, xia_sheared_2024}.

All evaluations were conducted using the \textbf{EleutherAI LM Evaluation Harness} (v0.4.9.1) \cite{lintang_sutawika_eleutherailm-evaluation-harness_2024}, a widely adopted framework that ensures reproducibility and comparability with other works. The prompts and few-shot configurations adhered to the framework's standard implementations.

Experiments were run on Google Colab with NVIDIA L4 GPUs (24GB VRAM). Inference employed $\text{torch.bfloat16}$ precision with the $\text{device\_map="auto"}$ loading strategy to optimize memory usage. The total evaluation time per model was approximately 5-6 hours, encompassing all benchmarks and pruning configurations.

\subsection{Reproducibility}
To ensure full reproducibility of our experiments, all pipeline components are publicly available and fully documented:

\begin{itemize}
    \item \textbf{Software and configurations:} We utilized \textbf{OptIFAIR} v0.2.0 \cite{martra_optipfair_2024} for structural pruning and $\text{lm-evaluation-harness}$ v0.4.9.1 \cite{lintang_sutawika_eleutherailm-evaluation-harness_2024} for evaluations. All pruning configurations, including the PPM method, applied percentages, and calibration seeds, are documented in the project's public repository. The benchmark evaluations adhered to the default harness configurations without modifications to the prompts.
    \item \textbf{Data and checkpoints:} The baseline models were sourced directly from HuggingFace Hub ($\text{meta-llama/Llama-3.2-1B}$ and $\text{meta-llama/Llama-3.2-3B}$) \cite{meta_ai_llama-modelsmodelsllama3_2model_cardmd_2024}. Complete results in JSON format, including comprehensive experimental metadata, are accessible in the project repository.
\item \textbf{Source code:} The entire codebase to reproduce the pruning process, evaluations, and all analyses is available 
\ifsubmission
\else
    on GitHub: \url{https://github.com/peremartra/llama-glu-expansion-pruning}. 
\fi
The repository contains documented notebooks for each experimental phase and scripts to regenerate all figures presented in the paper.
\end{itemize}

\section{Results}
\label{sec:Results}
This section presents the empirical findings from applying our PPM-based width pruning methodology (\ref{sec:methodology}) to the Llama-3.2-1B and 3B models. We analyze the impact of reducing the GLU expansion ratio using a comprehensive suite of capacity and efficiency benchmarks \cite{muralidharan_compact_2024}.

The results support our hypothesis of selective capacity degradation. We find that pruning does not induce uniform degradation but rather a set of systematic trade-offs between different task types and operational efficiency metrics (see Section \ref{sec:paradox}).

\subsection{Overall Performance Landscape}
\label{sec:performance}
We evaluated seven expansion ratio configurations across two Llama-3.2 models (1B and 3B parameters) using 13 benchmarks covering factual knowledge, mathematical and algorithmic reasoning, language understanding, truthfulness, and instruction-following. The results reveal markedly heterogeneous degradation patterns across different types of knowledge. Table 3 presents a summary of the most representative benchmarks at key expansion ratios: baseline, the $2.4\times$ ratio (identified as the equilibrium point in both models), and the most aggressive ratio evaluated. The following subsections systematically characterize three main findings:

\begin{itemize}
    \item A dichotomy between fragile and robust capabilities (see Section \ref{sec:dicothomy})
    \item An inverse correlation between knowledge and TruthfulQA-MC2 (see Section \ref{sec:paradox})
    \item Trade-offs between energy efficiency and inference latency (see Section \ref{sec:efficiency})
\end{itemize}

Complete results for all expansion ratios and benchmarks are available in Appendix \ref{sec:completebenchmarks}. 

\begin{table}[H]
\caption{Performance Summary at Key Expansion Ratios. Values in bold indicate performance improvements relative to baseline.}
\label{tab:performance-summary}
\centering
\vspace{0.3cm}
\begin{tabular}{l|ccc|ccc}
\toprule
& \multicolumn{3}{c|}{\textbf{Llama-3.2-1B}} & \multicolumn{3}{c}{\textbf{Llama-3.2-3B}} \\
\textbf{Benchmark Category} & \textbf{4.0×} & \textbf{2.4×} & \textbf{1.6×} & \textbf{2.67×} & \textbf{2.4×} & \textbf{1.07×} \\
\midrule
MMLU (Knowledge) & 0.311 & 0.269 & 0.255 & 0.561 & 0.433 & 0.259 \\
GSM8K (Math Reasoning) & 0.066 & 0.021 & 0.021 & 0.268 & 0.142 & 0.023 \\
IFEval (Instruction Following) & 0.104 & \textbf{0.152} & \textbf{0.137} & 0.094 & \textbf{0.131} & \textbf{0.133} \\
MUSR (Algorithmic Reasoning) & 0.340 & \textbf{0.429} & \textbf{0.409} & 0.364 & \textbf{0.373} & 0.360 \\
TruthfulQA-MC2 (Truthfulness) & 0.377 & \textbf{0.430} & \textbf{0.466} & 0.392 & 0.377 & \textbf{0.457} \\
\bottomrule
\end{tabular}
\end{table}

Complete results for all 7 expansion ratios and 13 benchmarks are provided in Appendix \ref{sec:completebenchmarks}.

\subsection{The Capability Dichotomy}
\label{sec:dicothomy}
The analysis of performance trajectories across expansion ratios reveals markedly contrasting patterns between different benchmarks. While some metrics show predictable and monotonic collapse with expansion ratio reduction, others exhibit unexpected behaviors: initial improvement followed by gradual degradation, or even sustained improvement across multiple pruning levels \cite{hou_instruction-following_2025}. Figures 2 and 3 illustrate these contrasting patterns for six representative benchmarks in Llama-3.2-1B and Llama-3.2-3B, respectively. These patterns reveal a systematic dichotomy that remains consistent between both model sizes under the PPM (Peak-to-Peak Magnitude) neuron selection method.

\begin{figure}[H]
\centering
\includegraphics[width=0.8\textwidth]{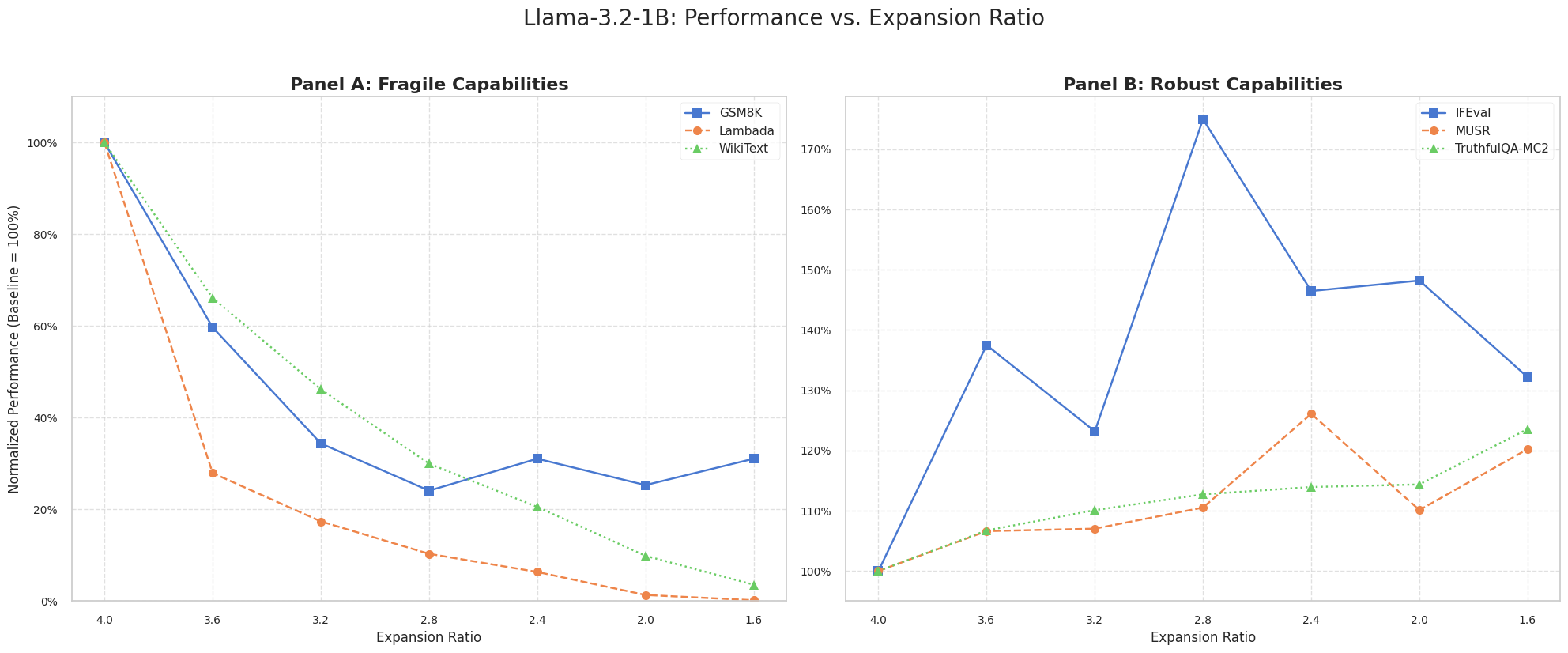}
\caption{Llama-3.2-1B Benchmarks. Panel A (Fragile Capabilities) shows the predictable collapse of knowledge-dependent tasks (GSM8K, Lambada, WikiText) as expansion ratio decreases. Panel B (Robust Capabilities) reveals the contrasting improvement of algorithmic and instruction-following tasks (IFEval, MUSR, TruthfulQA-MC2). Performance is normalized to baseline (4.0×) = 100\%. The X-axis represents expansion ratios from 4.0× to 1.6×; the Y-axis shows normalized performance.}
\label{fig:benchmarks-llama1b}
\end{figure}

\begin{figure}[H]
\centering
\includegraphics[width=0.8\textwidth, 
                       clip, 
                       trim={0cm 0cm 0cm 0cm}]{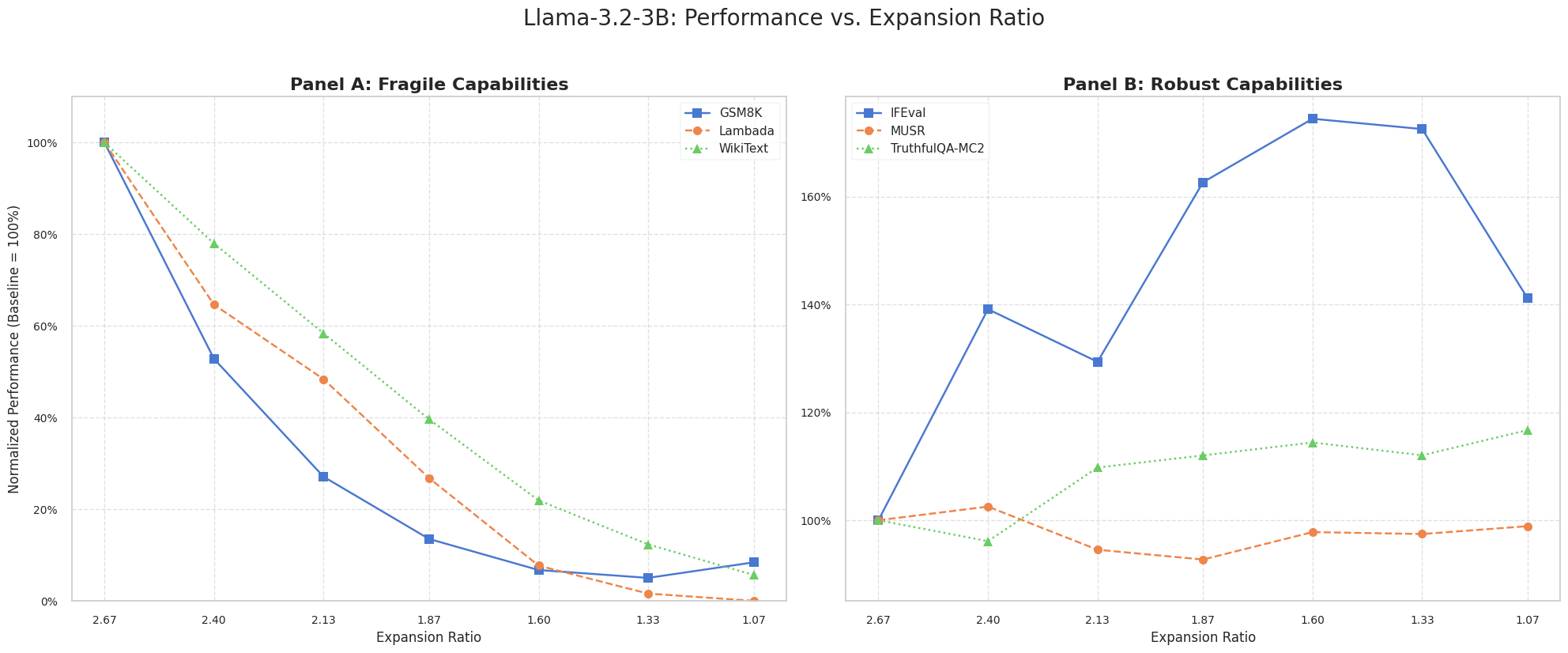}]
\caption{Llama-3.2-3B Benchmarks. Panel A (Fragile Capabilities) demonstrates near-monotonic degradation of knowledge-dependent tasks (GSM8K, Lambada, WikiText) across expansion ratios 2.67× to 1.07×. Panel B (Robust Capabilities) exhibits non-monotonic improvement in instruction-following (IFEval reaching 174.4\% of baseline) and rising TruthfulQA-MC2 scores. Performance is normalized to baseline (2.67×) = 100\%. This pattern replicates the dichotomy observed in Llama-1B.}
\label{fig:benchmarks-llama3b}
\end{figure}

These diverging patterns reveal two distinct categories of cognitive capabilities. \textbf{Fragile capabilities} (Panel A in both figures) include tasks that critically depend on parameterized knowledge stored in the model's weights \cite{sharma_truth_2023}. Mathematical reasoning (GSM8K) illustrates this behavior in a particularly dramatic way: in Llama-1B, accuracy collapses to $59.7\%$ of baseline at an expansion ratio of $3.6\times$ (after the first $10\%$ reduction), dropping precipitously to $31.1\%$ at $2.4\times$ and stabilizing around $31.1\%$ at $1.6\times$. Llama-3B exhibits a similar trajectory: $52.8\%$ of baseline at $2.4\times$ ($0.142$), declining to a minimum of $0.014$ at $1.33\times$ before a small rebound to $0.023$ at the most aggressive ratio ($1.07\times$). Perplexity metrics show even more severe collapses: Lambada degrades exponentially in both models, with WikiText demonstrating greater relative resilience but eventually converging toward the same fate at extreme ratios \cite{muralidharan_compact_2024} (see Appendix \ref{sec:completebenchmarks}). 

We note, however, that GSM8K accuracy in Llama-1B remains near the floor throughout (baseline $0.066$), so its trajectory is noisy and non-monotonic at the lower expansion ratios; we therefore treat the 1B GSM8K curve as illustrative, with the cleaner near-monotonic collapse of Llama-3B (from a baseline of $0.268$ down to a minimum of $0.014$, with a small floor-level rebound to $0.023$ at the most aggressive ratio) carrying the primary evidential weight for this benchmark.

In marked contrast, \textbf{robust capabilities} (Panel B) encompass tasks that require algorithmic processing or behavioral adherence rather than factual knowledge retrieval. Instruction-following (IFEval) improves with moderate pruning in both models (see Appendix \ref{sec:completebenchmarks}), reaching peaks of $175.0\%$ of baseline in Llama-1B (expansion ratio $2.8\times$) and $174.4\%$ in Llama-3B (expansion ratio $1.6\times$). Notably, even at more aggressive pruning levels, IFEval remains significantly above baseline: $132.2\%$ in Llama-1B ($1.6\times$) and $141.1\%$ in Llama-3B ($1.07\times$). Algorithmic reasoning (MUSR) shows differentiated patterns between models: Llama-1B reaches its maximum of $126.1\%$ of baseline at an expansion ratio of $2.4\times$, while Llama-3B stays close to baseline with only moderate variations. TruthfulQA-MC2 exhibits gradual and sustained improvement in both models, reaching $123.6\%$ of baseline in Llama-1B ($1.6\times$) and $116.7\%$ in Llama-3B ($1.07\times$), a pattern explored in detail in Section \ref{sec:paradox}.

This dichotomy, consistent across both model sizes, suggests that width pruning guided by the PPM importance criterion selectively modifies different functional components of the model. Fragile capabilities appear to critically depend on neurons that the PPM criterion prioritizes for elimination, as evidenced by their near-monotonic degradation as the expansion ratio decreases. Robust capabilities, by contrast, seem to depend more on processing patterns distributed across the entire transformer architecture, where dimensionality reduction in MLP layers can even act as a form of regularization that improves generalization by reducing overfitting to spurious patterns memorized during pre-training \cite{sharma_truth_2023}. This interpretation is supported by the particularly pronounced behavior of IFEval, whose improvement suggests that strict instruction adherence benefits from the elimination of spurious correlations in MLP layers.

\subsection{An Inverse MMLU–TruthfulQA-MC2 Correlation}
\label{sec:paradox}
A further pattern in our results is an inverse relationship between factual knowledge and the TruthfulQA-MC2 score: as MMLU degrades with pruning, MC2 rises (see Appendix~\ref{sec:completebenchmarks}). This pattern is observed consistently across both evaluated models. As we show below, it is most plausibly explained as a scoring effect of MC2 under degradation rather than as a gain in truthfulness.

\begin{figure}[h!]
\centering
\includegraphics[width=0.8\textwidth, 
                       clip, 
                       trim={0cm 0cm 0cm 0cm}]{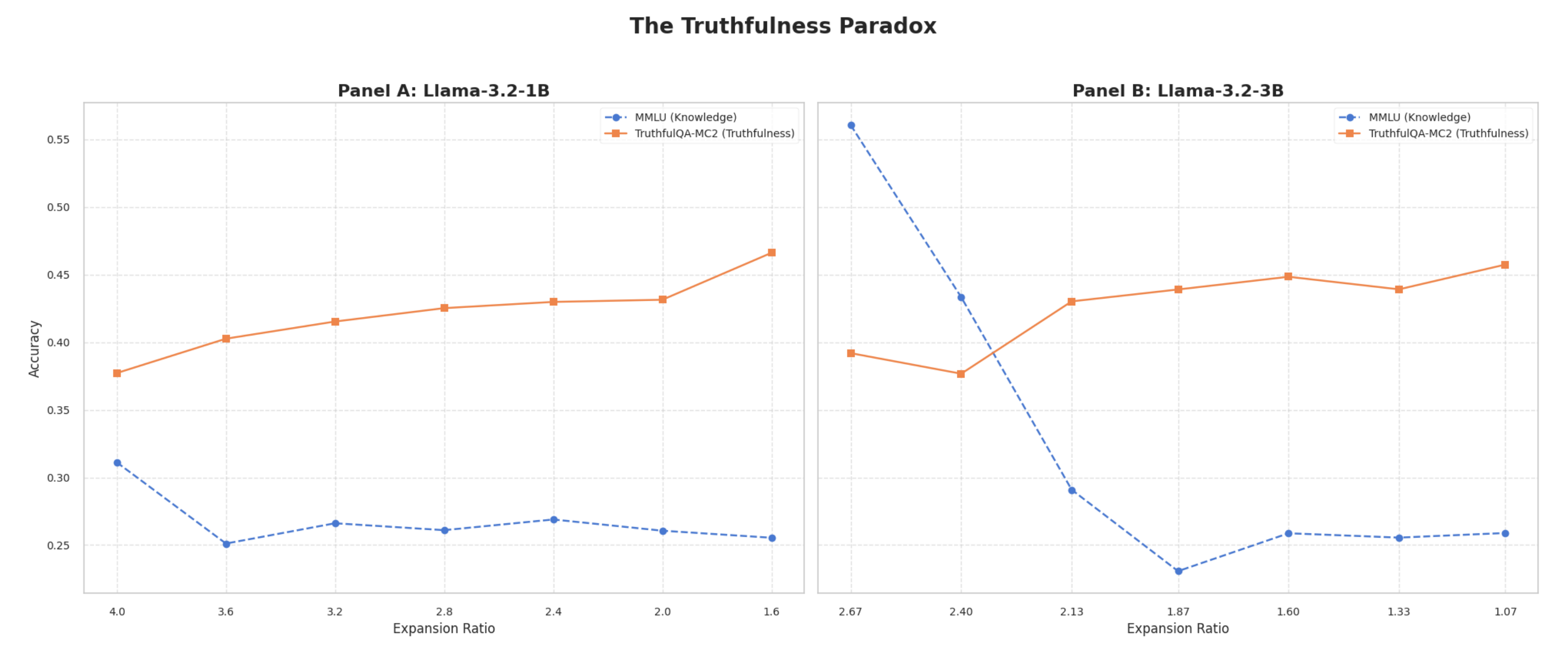}]
\caption{Inverse MMLU–TruthfulQA-MC2 Correlation. Divergent trajectories of factual knowledge (MMLU, blue dashed lines) and TruthfulQA-MC2 scores (orange solid lines) across expansion ratios. Panel A shows Llama-3.2-1B; Panel B shows Llama-3.2-3B. As the expansion ratio decreases, MMLU declines while TruthfulQA-MC2 rises, revealing a systematic inverse relationship quantified by the correlation analysis in Section~\ref{sec:paradox}.}
\label{fig:truthfulness-paradox}
\end{figure}

Figure~\ref{fig:truthfulness-paradox} illustrates this inverse relationship: MMLU performance degrades monotonically until it stabilizes near a floor (blue dashed lines), while TruthfulQA-MC2 scores rise (orange solid lines) across the evaluated expansion ratios. In Llama-1B (Panel A), MMLU accuracy declines from $0.311$ (baseline) to $0.255$ ($1.6\times$, $-17.9\%$), while TruthfulQA-MC2 rises from $0.377$ to $0.466$ ($+23.6\%$). In Llama-3B (Panel B), the pattern is more pronounced: MMLU accuracy declines from $0.561$ (baseline) to $0.259$ ($1.07\times$, $-53.8\%$), while TruthfulQA-MC2 rises from $0.392$ to $0.457$ ($+16.7\%$). Correlation analysis quantifies the relationship: Pearson $r = -0.864$ ($p = 0.012$) for Llama-3B. For Llama-1B, the trend is consistent in direction ($r = -0.676$) but does not reach statistical significance ($p = 0.096$), given the limited number of observations ($n = 7$).

PPM-based width pruning yields an observable association: as MMLU degrades, TruthfulQA-MC2 scores rise. TruthfulQA-MC2 scores a model over a distribution of true and false options, so a degraded model with flatter output distributions can raise its MC2 score without any gain in discriminating truth from falsehood. Our data are consistent with this: while MC2 rises, TruthfulQA-MC1---which rewards selecting the single correct answer---stays essentially flat ($0.234$ to $0.238$ in Llama-1B), whereas a genuine improvement would be expected to lift MC1 as well. We therefore interpret the inverse MMLU--MC2 correlation primarily as a measurement artifact of MC2 under degradation.

A complete characterization of this balance between different cognitive capabilities informs the analysis of optimal configurations, which we further develop in Section \ref{sec:discussions}.

\subsection{Efficiency Trade-offs: Single-Request vs. Batch Processing}
\label{sec:efficiency}
PPM-based width pruning yields energy efficiency improvements \cite{gholami_can_2023} that vary significantly depending on the inference mode. In a \textbf{Single-Request} configuration ($\text{batch\_size}=1$), Llama-1B reduces energy consumption from $0.268$ J/token (baseline, $4.0\times$) to $0.222$ J/token at $2.4\times$ ($-17.2\%$) and to $0.206$ J/token at $1.6\times$ ($-23.1\%$), as measured using MMLU as a representative benchmark for long-generation tasks. However, this improvement comes at the cost of increased end-to-end latency in aggressive pruning configurations. Although latency exhibits variability across intermediate expansion ratios, the general trend in aggressive configurations shows an increase: it reaches $929$ ms at $2.4\times$ ($+12.7\%$ compared to baseline) and $970$ ms at $1.6\times$ ($+17.7\%$), highlighting a trade-off between energy efficiency and response latency (see Appendix \ref{sec:energy}).

Figure 5 illustrates this trade-off and demonstrates that the deployment context critically determines its practical relevance. In a \textbf{Batch Processing} configuration ($\text{batch\_size}=8$), energy efficiency improves dramatically: Llama-1B consumes only $0.055$ J/token at baseline ($79.5\%$ lower than single-request) and $0.048$ J/token at $2.4\times$ ($78.3\%$ lower than single-request).

Pruning benefits both configurations, but the absolute difference between inference modes persists: the pruned model at $2.4\times$ consumes $0.222$ J/token in single-request mode versus $0.048$ J/token in batch processing---a $4.6\times$ difference. Llama-3B follows similar patterns, with comparable advantages in batch processing.

\begin{figure}[htbp]
  \centering
  \includegraphics[width=0.9\textwidth]{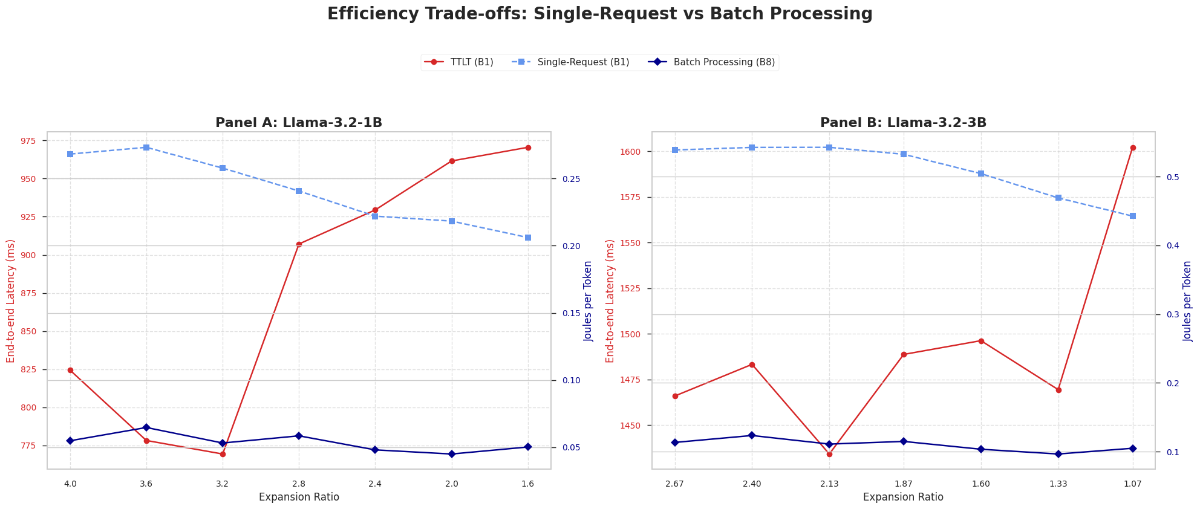}
  \caption{Efficiency Trade-offs --- Single-Request vs Batch Processing. Panel A (Llama-3.2-1B) and Panel B (Llama-3.2-3B) display the relationship between end-to-end latency (red lines), single-request energy consumption (blue dashed lines), and batch processing energy consumption (blue solid lines) across expansion ratios. Latency increases with pruning intensity in both models. Single-request energy decreases across expansion ratios, while batch processing energy remains consistently lower than single-request configurations at all expansion ratios. The left Y-axis shows latency in milliseconds; the right Y-axis shows energy consumption in joules per token.}
  \label{fig:efficiency_tradeoffs}
\end{figure}
This pattern suggests that pruned configurations are better optimized for batch processing workloads (see Section \ref{sec:efficiency}; see Appendix \ref{sec:energy}) than for interactive, single-request applications. In high-concurrency scenarios---where multiple requests are processed simultaneously---the fixed cost of prompt processing is amortized across parallel generations, effectively diluting the impact of increased individual end-to-end latency on overall system throughput. 

\section{Discussions}
\label{sec:discussions}
This section interprets the empirical findings presented in Section \ref{sec:Results}. We begin by analyzing the theoretical implications of the capacity dichotomy (Section \ref{sec:interpretationdico}) and the inverse MMLU–TruthfulQA-MC2 correlation (Section \ref{sec:interpretationpara}), then derive practical consequences for deployment (Section \ref{sec:practical}), and finally contextualize the work by addressing its limitations (Section \ref{sec:limitations}) and future directions (Section \ref{sec:future}).

\subsection{Interpretation of the Capacity Dichotomy}
\label{sec:interpretationdico}
The systematic dichotomy between fragile capabilities (e.g., GSM8K, MMLU, perplexity) and robust capabilities (e.g., IFEval, MUSR, TruthfulQA-MC2), as observed in Section \ref{sec:dicothomy}, remains consistent across both Llama-3.2-1B and Llama-3.2-3B. This consistency suggests that the dichotomy is not an artifact specific to a single model but rather a reproducible architectural pattern. We propose that the peak-to-peak magnitude of weights---the criterion employed by the PPM method---reflects distinct computational roles within the GLU-MLP layers: neurons with high-weight magnitudes may be associated with algorithmic processing and adherence to instructional structures (e.g., IFEval, MUSR) (see Section \ref{sec:dicothomy}), whereas neurons with lower-weight magnitudes primarily contribute to the storage and retrieval of parameterized factual knowledge (e.g., MMLU, GSM8K). 

This interpretation is supported by the differential severity of performance degradation: while GSM8K accuracy degrades markedly in Llama-3B (from a $0.268$ baseline down to a minimum of $0.014$ at $1.33\times$, a near-monotonic collapse), the Llama-1B drop, although larger in relative terms, starts from a near-floor $0.066$ baseline and is therefore treated as illustrative rather than as primary evidence. In contrast, instruction-following metrics exhibit substantial improvements ($+46.5\%$ and $+39.1\%$, respectively) (see Appendix~\ref{sec:completebenchmarks}).

Our finding that static PPM pruning improves instruction-following performance aligns with the work of Hou et al. \cite{hou_instruction-following_2025}, who also aim to enhance instruction-following through pruning. However, their approach relies on a dynamic method based on instruction activations, which contrasts with our static, weight-based strategy.

The specificity of the neuron selection method is critical to achieving 
this behavior. A preliminary evaluation comparing PPM with three 
alternative weight-only criteria (the standard L2 norm, VOW, and PON) 
revealed that all three lead to catastrophic performance collapse at a 
$10\%$ pruning ratio in Llama-3.2-1B: each drove LAMBADA perplexity 
into the thousands of percent above baseline, with L2 and PON 
additionally causing WikiText perplexity to diverge, while PPM produced 
moderate increases ($+54\%$ in WikiText, $+230\%$ in LAMBADA) 
(Appendix~\ref{sec:ppmjustification}). This evidence indicates that the 
importance criterion strongly influences which task capabilities are 
preserved during pruning.

Additionally, recent activation-based width pruning methods---such as SliceGPT \cite{ashkboos_slicegpt_2024}, which applies PCA to activation covariance matrices, and AMP \cite{mugnaini_efficient_2025}, which uses projected activation magnitude---do not report similar dichotomies. Instead, these methods focus on minimizing uniform performance degradation across tasks. This divergence suggests that the observed dichotomy is specific to the PPM criterion and not a general property of width pruning in GLU architectures.

These observations carry direct implications for the design of pruning strategies. Importance criteria are not interchangeable; different methods selectively impact distinct cognitive capabilities of the model. While activation-based approaches prioritize preserving uniform performance, PPM introduces a selective trade-off that may be advantageous in applications where instructional adherence take precedence over the exhaustiveness of factual knowledge.

\subsection{Interpretation of the Inverse MMLU–TruthfulQA-MC2 Correlation}
\label{sec:interpretationpara}
The inverse correlation between MMLU and TruthfulQA-MC2 documented in Section~\ref{sec:paradox} has a straightforward reading. The rise in MC2 (from $0.377$ to $0.466$ in Llama-1B, $+23.6\%$) is accompanied by near-flat MC1 (from $0.234$ to $0.238$, $+1.6\%$). Since MC1 rewards selecting the single correct answer while MC2 scores the full distribution over options, this MC1-flat / MC2-rising pattern is what we would expect from a degraded model whose output distribution flattens.

A plausible reading is that pruning lowers the model's confidence, flattening its output distribution; MC2 rewards this flattening as a scoring side effect.

The practical takeaway is narrower and methodological: TruthfulQA-MC2 should be interpreted with care as a standalone metric under compression, since it can rise for reasons unrelated to genuine gains in truthfulness. We do not rule out a real improvement in truthfulness, but our data do not allow us to separate it from this scoring effect.

\subsection{Practical Implications: Balancing Capabilities and Efficiency}
\label{sec:practical}
Energy efficiency improves consistently up to an expansion ratio of $2.4\times$, but this improvement comes with critical trade-offs that depend on the inference mode \cite{gholami_can_2023}. In a \textbf{single-request} configuration ($\text{batch\_size}=1$), Llama-1B reduces energy consumption by $17.2\%$ (from $0.268$ to $0.222$ J/token) but increases end-to-end latency by $12.7\%$ (from $824$ ms to $929$ ms). In contrast, \textbf{batch processing} ($\text{batch\_size}=8$) offers substantially superior efficiency ($0.048$ J/token, approximately $4.6\times$ better than single-request mode) with throughput resilience. These results suggest distinct optimization profiles for different deployment scenarios (see Section \ref{sec:efficiency}).

It is worth noting that the end-to-end latency analysis reveals non-monotonic behavior at intermediate expansion ratios (a decrease between $3.6\times$ and $3.2\times$, followed by a sustained increase), possibly due to hardware optimizations or cache memory effects. However, this pattern does not invalidate the general trend of increasing latency in more aggressive pruning configurations.

The $2.4\times$ expansion ratio emerges as an equilibrium point for balancing capabilities in the two evaluated models. However, this convergence should be interpreted cautiously: with only two model sizes (1B and 3B), we cannot claim that $2.4\times$ is universally optimal for the entire Llama family or for GLU architectures in general. Notably, achieving a $2.4\times$ ratio requires substantially different pruning percentages in each model: $40\%$ in Llama-1B (from a baseline of $4.0\times$) versus only $10\%$ in Llama-3B (from a baseline of $2.67\times$). This indicates that the equilibrium expansion ratio is not a function of the pruning percentage but rather a property of the resulting architectural ratio.

At this ratio, both models retain robust algorithmic capabilities ($\text{IFEval}$: $+46.5\%$ and $+39.1\%$, respectively) (see Appendix \ref{sec:completebenchmarks}) while maintaining factual knowledge at levels acceptable for many applications ($\text{MMLU}$: $86.4\%$ and $77.3\%$ of baseline). However, mathematical reasoning capabilities exhibit severe degradation ($\text{GSM8K}$: $31.1\%$ in Llama-1B, $52.8\%$ in Llama-3B). This reveals a scale effect where larger models demonstrate greater fragility but also greater relative resilience in fragile tasks.

It is critical to recognize that the $2.4\times$ ratio constitutes an equilibrium only under specific application priorities. For applications that prioritize the exhaustiveness of factual knowledge (e.g., $\text{MMLU}$, $\text{GSM8K}$) over instructional adherence, higher expansion ratios---or even the unpruned baseline---would be more appropriate. The "optimal point" fundamentally depends on the application's objectives \cite{reda_how_2025}.

The practical guidelines derived from these results emphasize the importance of deployment context:
\begin{itemize}
    \item \textbf{Batch processing workloads} (e.g., offline generation, document analysis) can leverage $2.4\times$ or more aggressive configurations to maximize energy efficiency (see Appendix \ref{sec:energy}).
    \item \textbf{Instruction-following oriented applications} with modest encyclopedic knowledge requirements may find the $2.4\times$ ratio to be an optimal balance between improved behavioral adherence and acceptable knowledge degradation (see Appendix \ref{sec:completebenchmarks}).
\end{itemize}

Unlike previous work on width pruning, such as \textbf{SliceGPT} and \textbf{AMP}---which does not report systematic exploration of optimal expansion ratios or deployment-specific trade-offs---our approach reveals that the optimal architectural ratio depends critically on both the neuron selection method and the specific use case \cite{mugnaini_efficient_2025, reda_how_2025}.

\subsection{Limitations}
\label{sec:limitations}
This study has several methodological and scope limitations that must be considered when interpreting the results. First, our analysis is limited to two model sizes within the Llama-3.2 family (1B and 3B parameters), both of which fall within the small model range. As a result, our conclusions about width pruning behavior using the PPM method cannot be extrapolated to larger models (7B, 13B, 70B+), where the distribution of capabilities and resilience to pruning may differ significantly.

Second, we exclusively evaluate GLU architectures as implemented in Llama-3.2. The observed capacity dichotomy and the $2.4\times$ equilibrium point may not generalize to other architectural families (e.g., Mistral, Qwen, Gemma) or to MLP variants without gating mechanisms \cite{guo_dependency-aware_2024, shazeer_glu_2020}.

Third, we employ only the PPM (Peak-to-Peak Magnitude) method for neuron selection, having empirically validated its superiority over VOW, PON and L2. However, we do not explore other potentially relevant criteria---such as gradient-based importance, second-order methods, or activation-aware techniques---that could reveal different trade-offs \cite{ai_nirvana_2025}. It is critical to emphasize that our main findings, particularly the dichotomy between fragile and robust capabilities and the inverse MMLU–TruthfulQA-MC2, are specific to the PPM criterion. As demonstrated in our preliminary experiments (Appendix \ref{sec:ppmjustification}), alternative methods like VOW, PON and L2 lead to catastrophic performance collapse even at minimal pruning levels. Therefore, these patterns should not be generalized to width pruning as a whole, but specifically to width pruning using PPM selection.

The experimental scope also presents additional limitations. The main results (Section \ref{sec:Results}) focus on base models without instruction tuning, with analysis of instruct models deferred to Appendix \ref{sec:analysisinstruct}. Since instruction tuning substantially alters the distribution of knowledge in the weights, we cannot confirm that the observed patterns persist in models fine-tuned for instruction following. Furthermore, we do not explore post-pruning recovery strategies through additional fine-tuning, which could potentially mitigate the degradation of fragile capabilities such as GSM8K \cite{muralidharan_compact_2024}. We also do not conduct controlled comparisons with alternative compression techniques---such as quantization (reducing numerical precision) or knowledge distillation (transferring to smaller architectures)---which might offer different trade-offs between efficiency and capabilities. Finally, we apply uniform pruning across all model layers and do not investigate whether layer-selective pruning (preserving specific early or late layers) or non-uniform pruning could improve the balance of capabilities \cite{sharma_truth_2023}.

These limitations do not invalidate the reported findings but do define their scope. The qualitative patterns observed---the dichotomy between fragile and robust capabilities \cite{wei_assessing_2024}, the inverse MMLU–TruthfulQA-MC2 correlation, and the trade-off between energy efficiency and end-to-end latency---are robust within the evaluated context (small Llama-3.2 models, PPM method, base configuration). However, they may generalize to broader contexts only with further empirical validation. We further note that GSM8K accuracy in Llama-1B remains near the floor across all configurations (baseline $0.066$), which limits the interpretability of relative changes on this benchmark for the 1B model; we therefore rely on the higher-baseline Llama-3B trajectory as the primary evidence for degradation of knowledge-dependent reasoning. In contrast, the specific equilibrium point at a $2.4\times$ expansion ratio is the most context-dependent finding: consistently observed in only two model sizes (1B and 3B), each with different architectural baselines ($4.0\times$ and $2.67\times$). Its transferability to larger models or distinct architectural families remains uncertain and requires specific investigation.

\subsection{Future Work}
\label{sec:future}
The findings of this study, together with the limitations identified in Section \ref{sec:limitations}, open multiple avenues for future research. The most immediate direction involves extending the analysis to larger models within the Llama family (7B, 13B, 70B) and other GLU-based architectures (e.g., Mistral, Qwen). This extension aims to determine whether the qualitative patterns observed persist:

\begin{itemize}
    \item The dichotomy between fragile capabilities (parameterized factual knowledge) and robust capabilities (algorithmic processing and instruction-following),
    \item The inverse MMLU–TruthfulQA-MC2 correlation, and whether it reflects a genuine change in truthfulness or, as our analysis suggests, a scoring effect of MC2 under degradation, and
    \item The systematic trade-off between energy efficiency and end-to-end latency, depending on the inference mode.
\end{itemize}

Additionally, it is essential to determine whether an analogous architectural equilibrium point---though not necessarily at $2.4\times$---exists in larger-scale models, balancing these competing tensions. Such findings would provide insights into the generality of our observations. The code repository accompanying this study includes a configurable pipeline designed to facilitate this extension, requiring only the specification of the target model and desired pruning ratios.

\textbf{Exploration of Alternative Importance Scoring Methods}

Investigating alternative importance scoring methods---particularly hybrid criteria that combine weight magnitude with activation statistics---could clarify whether the observed capability dichotomy is specific to the PPM method \cite{ai_nirvana_2025} or a more fundamental property of width pruning in GLU architectures. This research direction also raises a broader hypothesis for future exploration: width pruning could serve not only as a compression technique but also as a tool for functional specialization and behavior modification in zero-shot settings \cite{sharma_truth_2023}.

We propose that the importance criterion acts as a control mechanism for this specialization. By designing custom criteria (e.g., activation-based), it may be possible to selectively shape the model's capability profile \cite{sharma_truth_2023}, enhancing specific domains or mitigating biases.

\textbf{Impact of PPM Pruning on Instruction-Tuned Models}

Further research should also examine how PPM pruning affects models that have undergone instruction tuning \cite{hou_instruction-following_2025}. Our preliminary analysis (Appendix \ref{sec:analysisinstruct}) reveals a counterintuitive finding: although the Llama-1B-Instruct model experiences a drop in IFEval performance from $36.41\%$ to $14.6\%$ after $40\%$ pruning (an apparent degradation of $-59.9\%$), this final performance level converges to that of the pruned base model in the same configuration ($15.16\%$).

This suggests that PPM pruning does not degrade the model's fundamental instruction-following capabilities. Instead, it selectively eliminates the specific improvements introduced by fine-tuning---the gains from $10.35\%$ to $36.41\%$ that instruction tuning had added to the base model. The most parsimonious interpretation is that instruction tuning stores its modifications in neurons with low peak-to-peak weight magnitudes---precisely those that PPM prioritizes for elimination---while base capabilities reside in high-magnitude neurons.

\section{Conclusion}
\label{sec:conclusion}
This study presents a systematic analysis of the impact of width pruning on GLU-MLP layers in Llama-3.2 models, specifically examining how variations in the expansion ratio affect different cognitive capabilities. Despite being a fundamental architectural parameter, the expansion ratio has received limited empirical attention in the structured pruning literature \cite{zhu_survey_2024}. Our results demonstrate that width pruning guided by the Peak-to-Peak Magnitude (PPM) criterion does not merely serve as uniform compression. Instead, it acts as a selective intervention that modifies distinct model functions, revealing systematic trade-offs between cognitive capabilities and operational efficiency metrics. These findings suggest that width pruning, through the deliberate design of importance criteria, could be employed as a tool for selective behavioral modification rather than simple compression \cite{sharma_truth_2023}---a direction we propose to explore in future work (Section \ref{sec:future}).

\textbf{Capability dichotomy.} The central finding of this study is the existence of a reproducible dichotomy between fragile and robust capabilities \cite{wei_assessing_2024}. While tasks reliant on parametric knowledge---such as MMLU, GSM8K, and perplexity---degrade predictably with reductions in the expansion ratio \cite{muralidharan_compact_2024} (see Section \ref{sec:dicothomy}), instruction-following metrics exhibit significant improvements ($\text{IFEval}$: $+46.5\%$ in Llama-1B and $+39.1\%$ in Llama-3B at an expansion ratio of $2.4\times$ for each model) (Appendix \ref{sec:completebenchmarks}). This reveals a systematic trade-off between factual memory and behavioral adherence. This pattern remains consistent across both evaluated models (1B and 3B parameters), suggesting that it is not an artifact specific to a single model size but rather a reproducible architectural phenomenon. Critically, this dichotomy is specific to the PPM criterion. As demonstrated in our preliminary experiments, alternative pruning methods---such as VOW and PON---result in catastrophic performance collapse even at minimal pruning levels.

\textbf{Inverse MMLU--TruthfulQA-MC2 correlation.} We observe an inverse correlation ($r = -0.864$, $p = 0.012$ in Llama-3B) between MMLU and TruthfulQA-MC2 (Section~\ref{sec:paradox}). As MMLU degrades with expansion ratio reduction, MC2 rises while TruthfulQA-MC1 stays essentially flat (see Appendix~\ref{sec:completebenchmarks}). This MC1-flat / MC2-rising pattern is most consistent with a scoring effect of MC2 under degradation rather than a genuine improvement in truthfulness, and we report it as a methodological caution for the use of TruthfulQA-MC2 in compression research.

\textbf{Efficiency trade-offs depend on the inference mode.} Pruning reduces energy consumption by up to $23\%$ (J/token) but increases end-to-end latency by up to $17.58\%$ in single-request configurations ($\text{batch\_size}=1$). Batch processing ($\text{batch\_size}=8$) achieves $4.6\times$ superior energy efficiency with stable throughput (see Section \ref{sec:efficiency}).

A $2.4\times$ expansion ratio emerges as an equilibrium point in the two evaluated models, balancing algorithmic capability improvements ($+46\%$ and $+39\%$ in $\text{IFEval}$) (Appendix \ref{sec:completebenchmarks}) with controlled factual knowledge degradation ($86.4\%$ and $77.3\%$ of baseline MMLU) \cite{muralidharan_compact_2024}. However, it is critical to recognize that this "optimal point" depends fundamentally on application priorities \cite{reda_how_2025}: for systems prioritizing knowledge exhaustiveness over instructional adherence, higher expansion ratios---or even the unpruned baseline---would be more appropriate.

In summary, this work establishes that the expansion ratio in GLU-MLP layers is not merely a compression hyperparameter but an architectural variable that determines systematic trade-offs between different types of cognitive capabilities. The selection of the optimal ratio depends critically on the importance criterion used (PPM vs. alternatives) \cite{ai_nirvana_2025}, the deployment context (batch vs. single-request), and application priorities (instructional adherence vs. factual knowledge). 

The pruned models and complete code to reproduce all experiments are 
\ifsubmission
\else
    on GitHub: \url{https://github.com/peremartra/llama-glu-expansion-pruning}. 
\fi

\bibliography{references}
\bibliographystyle{tmlr}

\appendix
\section{Complete Benchmark Results (Base Models)}
\label{sec:completebenchmarks}
This appendix provides the complete results of the 13 benchmarks from the evaluation suite (described in Table 2) for all expansion ratio configurations of the Llama-3.2-1B and Llama-3.2-3B base models.

\textbf{Metric}: All scores are Accuracy or Acc-Norm (higher is better), except WikiText and Lambada, which are Perplexity (lower is better).

\textbf{Source}: Data is extracted from the project results files $\text{llama\_1b\_complete\_results\_latest.json}$ and $\text{llama\_3b\_complete\_results\_latest.json}$.

\begin{table}[H]
\caption{Complete Benchmark Results vs. Expansion Ratio (Llama-3.2-1B).}
\label{tab:results-1b}
\centering
\resizebox{\textwidth}{!}{%
\begin{tabular}{llccccccc}
\toprule
 &  & \textbf{4.0x} & \textbf{3.6x} & \textbf{3.2x} & \textbf{2.8x} & \textbf{2.4x} & \textbf{2.0x} & \textbf{1.6x} \\
\textbf{Category} & \textbf{Benchmark} & \textbf{(Base)} & \textbf{(10\%)} & \textbf{(20\%)} & \textbf{(30\%)} & \textbf{(40\%)} & \textbf{(50\%)} & \textbf{(60\%)} \\
\midrule
Knowledge & MMLU & 0.3111 & 0.2511 & 0.2661 & 0.2610 & 0.2689 & 0.2606 & 0.2554 \\
          & ARC-Challenge & 0.3626 & 0.3328 & 0.3080 & 0.2637 & 0.2509 & 0.2474 & 0.2398 \\
\addlinespace
Reasoning & GSM8K & 0.0660 & 0.0394 & 0.0227 & 0.0159 & 0.0205 & 0.0167 & 0.0205 \\
          & MUSR & 0.3399 & 0.3624 & 0.3638 & 0.3757 & 0.4286 & 0.3743 & 0.4087 \\
\addlinespace
Understanding & HellaSwag & 0.6363 & 0.5791 & 0.5076 & 0.4382 & 0.3737 & 0.3251 & 0.2909 \\
              & WinoGrande & 0.5991 & 0.6093 & 0.5935 & 0.5722 & 0.5706 & 0.5312 & 0.4870 \\
              & PIQA & 0.7454 & 0.7280 & 0.6757 & 0.6458 & 0.6115 & 0.5903 & 0.5637 \\
              & BoolQ & 0.6343 & 0.6260 & 0.6232 & 0.6260 & 0.6220 & 0.6141 & 0.5535 \\
\addlinespace
PPL & WikiText & 11.57 & 17.50 & 25.05 & 38.58 & 56.33 & 117.04 & 322.95 \\
(Lower is better) & Lambada & 5.75 & 20.59 & 33.07 & 55.74 & 90.38 & 428.30 & 2941.08 \\
\addlinespace
Truthfulness & TruthfulQA-MC1 & 0.2338 & 0.2460 & 0.2424 & 0.2448 & 0.2485 & 0.2460 & 0.2375 \\
             & TruthfulQA-MC2 & 0.3772 & 0.4026 & 0.4153 & 0.4252 & 0.4298 & 0.4314 & 0.4661 \\
\addlinespace
Instructions & IFEval & 0.1035 & 0.1423 & 0.1275 & 0.1811 & 0.1516 & 0.1534 & 0.1368 \\
\bottomrule
\end{tabular}%
}
\end{table}

\begin{table}[H]
\caption{Complete Benchmark Results vs. Expansion Ratio (Llama-3.2-3B).}
\label{tab:results-3b}
\centering
\resizebox{\textwidth}{!}{%
\begin{tabular}{llccccccc}
\toprule
 &  & \textbf{2.67x} & \textbf{2.4x} & \textbf{2.13x} & \textbf{1.87x} & \textbf{1.6x} & \textbf{1.33x} & \textbf{1.07x} \\
\textbf{Category} & \textbf{Benchmark} & \textbf{(Base)} & \textbf{(10\%)} & \textbf{(20\%)} & \textbf{(30\%)} & \textbf{(40\%)} & \textbf{(50\%)} & \textbf{(60\%)} \\
\midrule
Knowledge & MMLU & 0.5605 & 0.4333 & 0.2909 & 0.2307 & 0.2587 & 0.2555 & 0.2589 \\
          & ARC-Challenge & 0.4582 & 0.3959 & 0.3669 & 0.3123 & 0.2654 & 0.2381 & 0.2150 \\
\addlinespace
Reasoning & GSM8K & 0.2684 & 0.1418 & 0.0728 & 0.0364 & 0.0182 & 0.0136 & 0.0227 \\
          & MUSR & 0.3638 & 0.3730 & 0.3439 & 0.3373 & 0.3558 & 0.3545 & 0.3598 \\
\addlinespace
Understanding & HellaSwag & 0.7357 & 0.6853 & 0.6158 & 0.5232 & 0.4145 & 0.3399 & 0.2959 \\
              & WinoGrande & 0.6953 & 0.6748 & 0.6385 & 0.5927 & 0.5572 & 0.4886 & 0.4815 \\
              & PIQA & 0.7748 & 0.7508 & 0.7307 & 0.6812 & 0.6474 & 0.6045 & 0.5539 \\
              & BoolQ & 0.7294 & 0.5046 & 0.3972 & 0.4269 & 0.4208 & 0.5119 & 0.5034 \\
\addlinespace
PPL & WikiText & 9.26 & 11.88 & 15.86 & 23.35 & 42.18 & 74.83 & 162.47 \\
(Lower is better) & Lambada & 3.95 & 6.11 & 8.16 & 14.72 & 51.02 & 240.72 & 5960.46 \\
\addlinespace
Truthfulness & TruthfulQA-MC1 & 0.2497 & 0.2203 & 0.2387 & 0.2607 & 0.2448 & 0.2472 & 0.2387 \\
             & TruthfulQA-MC2 & 0.3919 & 0.3767 & 0.4302 & 0.4390 & 0.4484 & 0.4391 & 0.4574 \\
\addlinespace
Instructions & IFEval & 0.0943 & 0.1312 & 0.1220 & 0.1534 & 0.1645 & 0.1627 & 0.1331 \\
\bottomrule
\end{tabular}%
}
\end{table}

\section{Analysis of Instruct-Tuned Models (Llama-3.2-1B-Instruct)}
\label{sec:analysisinstruct}
The main analysis of this study (Section \ref{sec:Results}) focused on base models (pre-trained) to isolate the impact of width pruning on the fundamental capabilities acquired during pre-training. However, to understand how PPM pruning affects models that have been fine-tuned for instruction following, we conducted a complementary evaluation using Llama-3.2-1B-Instruct. This analysis is based on a single instruct-tuned model; we did not evaluate the 3B-Instruct model due to computational constraints.

The results reveal a critical finding: the "Capability Dichotomy" observed in base models does not replicate in the evaluated instruct-tuned model. Specifically, instruction-following capability ($\text{IFEval}$), which improved under moderate pruning in the base model ($+3.3$ percentage points from baseline to $1.6\times$), experiences severe degradation in the instruct-tuned model ($-22.7$ percentage points) (Appendix \ref{sec:analysisinstruct}), converging toward base model performance levels. In contrast, capabilities such as $\text{TruthfulQA-MC2}$ and $\text{MUSR}$ remain robust across both models, replicating the pattern observed in Section \ref{sec:paradox}.

We observed that, under PPM pruning, the performance of the instruct-tuned model converges toward that of the base model at low expansion ratios. Table B.1 compares the performance of $\text{Llama-3.2-1B (Base)}$ and $\text{Llama-3.2-1B-Instruct}$ at equivalent expansion ratios, highlighting fragile capabilities ($\text{MMLU}$, $\text{GSM8K}$, $\text{IFEval}$) alongside a robust capability ($\text{TruthfulQA-MC2}$) to illustrate this phenomenon.

\begin{table}[H]
\caption{Performance Comparison: Base vs. Instruct at Equivalent Expansion Ratios.}
\label{tab:base-vs-instruct}
\centering
\begin{tabular}{llcccc}
\toprule
 & \textbf{Exp.} & \textbf{MMLU} & \textbf{GSM8K} & \textbf{IFEval} & \textbf{TruthQA-MC2} \\
\textbf{Model} & \textbf{Ratio} & \small{(Knowledge)} & \small{(Reasoning)} & \small{(Instruction)} & \small{(Truthfulness)} \\
\midrule
Llama-1B (Base)     & 4.0x (0\%)  & 0.311 & 0.066 & 0.104 & 0.377 \\
Llama-1B-Instruct   & 4.0x (0\%)  & 0.456 & 0.339 & 0.364 & 0.438 \\
\addlinespace
Llama-1B (Base)     & 2.4x (40\%) & 0.269 & 0.021 & 0.152 & 0.430 \\
Llama-1B-Instruct   & 2.4x (40\%) & 0.261 & 0.017 & 0.146 & 0.437 \\
\addlinespace
Llama-1B (Base)     & 1.6x (60\%) & 0.255 & 0.021 & 0.137 & 0.466 \\
Llama-1B-Instruct   & 1.6x (60\%) & 0.246 & 0.019 & 0.137 & 0.444 \\
\bottomrule
\end{tabular}
\end{table}

As shown in the table, while baseline models ($4.0\times$) exhibit significant performance gaps between the Base and Instruct versions (e.g., $+26.0$ percentage points in $\text{IFEval}$, $+27.5$ percentage points in $\text{GSM8K}$), models pruned at $2.4\times$ and $1.6\times$ converge to nearly identical values for fragile capabilities. At $1.6\times$, $\text{IFEval}$ reaches exactly $0.137$ in both models (mathematical convergence), $\text{GSM8K}$ differs by only $0.2$ percentage points, and $\text{MMLU}$ by $0.9$ percentage points. Notably, $\text{TruthfulQA-MC2}$ does not show this convergence pattern and remains stable across both models at all expansion ratios.

This convergence pattern suggests that the PPM criterion, which prioritizes neurons with the highest peak-to-peak weight range, eliminates neurons whose removal results in the loss of capabilities added during instruction tuning, while preserving the base instruction-following capabilities inherent in the base model. The instruct-tuned model does not fall below the base model’s $\text{IFEval}$ performance (both converge to $13.7\%$), confirming that PPM does not eliminate base instruction-following capabilities but only the improvements introduced during fine-tuning.

This reinforces the hypothesis presented in Section \ref{sec:future}: the neural importance criterion acts as a control mechanism that determines which functional capabilities are preserved or removed. In the case of the evaluated instruct-tuned model, PPM selectively eliminates both knowledge-intensive capabilities ($\text{MMLU}$, $\text{GSM8K}$) \cite{muralidharan_compact_2024} and alignment improvements ($\text{IFEval}$) introduced by fine-tuning, while consistently preserving robust capabilities ($\text{TruthfulQA-MC2}$, $\text{MUSR}$), in line with the behavior observed in base models. These results are preliminary and specific to the PPM criterion as applied to $\text{Llama-3.2-1B-Instruct}$; further evaluations with additional instruct-tuned models would be necessary to generalize these findings.

\section{Complete Energy Efficiency Measurements}
\label{sec:energy}
Section \ref{sec:efficiency} presents the energy efficiency trade-off analysis between inference modes (batch size 1 vs. 8). Tables C.1 and C.2 provide the complete measurements for all evaluated expansion ratios in both models, including energy consumption (J/token), end-to-end latency, and throughput \cite{benoit_courty_mlco2codecarbon_2025}.

\begin{table}[H]
\caption{Llama-3.2-1B Energy Efficiency Metrics.}
\label{tab:energy-1b}
\centering
\resizebox{\textwidth}{!}{%
\begin{tabular}{ccccccc}
\toprule
& & \multicolumn{3}{c}{\textbf{Batch Size = 1 (Single Request)}} & \multicolumn{2}{c}{\textbf{Batch Size = 8 (Batch)}} \\
\cmidrule(lr){3-5} \cmidrule(lr){6-7} 
\textbf{Expansion} & \textbf{Pruning} & & \textbf{Latency} & \textbf{Throughput} & & \textbf{Throughput} \\
\textbf{Ratio} & \textbf{(\%)} & \textbf{J/token} & \textbf{(ms)} & \textbf{(tok/s)} & \textbf{J/token} & \textbf{(tok/s)} \\
\midrule
4.0x & 0\%  & 0.2767 & 877.30  & 50.90 & 0.0596 & 264.41 \\
3.6x & 10\% & 0.2763 & 1006.68 & 50.80 & 0.0696 & 262.95 \\
3.2x & 20\% & 0.2708 & 1123.24 & 51.15 & 0.0565 & 267.47 \\
2.8x & 30\% & 0.2529 & 1312.35 & 50.96 & 0.0618 & 265.49 \\
2.4x & 40\% & 0.2367 & 1320.89 & 50.80 & 0.0507 & 268.56$^{\star}$ \\
2.0x & 50\% & 0.2313 & 1400.91 & 51.48 & 0.0473 & 275.04 \\
1.6x & 60\% & 0.2179 & 1396.79 & 51.33 & 0.0525 & 273.43 \\
\bottomrule
\end{tabular}%
}
\end{table}

\begin{table}[H]
\caption{Llama-3.2-3B Energy Efficiency Metrics.}
\label{tab:energy-3b}
\centering
\resizebox{\textwidth}{!}{%
\begin{tabular}{ccccccc}
\toprule
& & \multicolumn{3}{c}{\textbf{Batch Size = 1 (Single Request)}} & \multicolumn{2}{c}{\textbf{Batch Size = 8 (Batch)}} \\
\cmidrule(lr){3-5} \cmidrule(lr){6-7}
\textbf{Expansion} & \textbf{Pruning} & & \textbf{Latency} & \textbf{Throughput} & & \textbf{Throughput} \\
\textbf{Ratio} & \textbf{(\%)} & \textbf{J/token} & \textbf{(ms)} & \textbf{(tok/s)} & \textbf{J/token} & \textbf{(tok/s)} \\
\midrule
2.67x & 0\%  & 0.5970 & 1267.70 & 29.51 & 0.1262 & 153.63 \\
2.40x & 10\% & 0.5851 & 1790.98 & 29.45 & 0.1340 & 145.64 \\
2.13x & 20\% & 0.5689 & 1893.44 & 29.81 & 0.1206 & 155.75 \\
1.87x & 30\% & 0.5691 & 2053.14 & 29.81 & 0.1238 & 155.29 \\
1.60x & 40\% & 0.5407 & 2031.00 & 29.94 & 0.1099 & 161.17 \\
1.33x & 50\% & 0.5025 & 2066.31 & 30.16 & 0.1019 & 164.35 \\
1.07x & 60\% & 0.4691 & 2391.20 & 30.04 & 0.1115 & 162.91 \\
\bottomrule
\end{tabular}%
}
\end{table}

\textbf{Notes:} J/token = joules per token (average energy per generated token); Latency = generation time per prompt (milliseconds); Throughput = tokens generated per second. Metrics represent averages across multiple benchmarks (HellaSwag, MMLU, IFEval) with three runs per configuration (seeds: 42, 123, 456).

\section{Justification for the Selection of the Importance Criterion (PPM)}
\label{sec:ppmjustification}
The methodology of this study (Section \ref{sec:methodology}) relies 
exclusively on the PPM (Peak-to-Peak Magnitude) criterion for neuron 
selection during pruning. This decision is based on a preliminary 
evaluation conducted prior to the main experiments (documented in the 
notebook $\text{00\_Neuron\_Selection\_Method\_Comparison.ipynb}$ in the 
project repository), which compared PPM against three additional 
weight-only importance criteria: L2 (sum of L2 norms), VOW (Variance of 
Weights), and PON (Product of L1 Norms).

The results demonstrate that the three alternatives are incompatible with 
gradual pruning in GLU architectures, leading to catastrophic collapse in 
perplexity metrics. At only $10\%$ pruning, all three drive LAMBADA 
perplexity into the thousands of percent above baseline, and L2 and PON 
additionally cause WikiText perplexity to diverge. PPM was the only 
criterion that allowed controlled degradation ($+230\%$ in LAMBADA, $+54\%$ 
in WikiText), enabling the gradual analysis reported in Sections 
\ref{sec:Results} and \ref{sec:discussions}.

\subsection*{Definition of the Evaluated Criteria}

All four criteria compared in this study are \emph{weight-only}: they 
rank neurons using only the values of the incoming weights, without 
reference to activations or gradients. For a GLU-MLP layer, each neuron 
$i$ is associated with a row $W^{\text{gate}}_i$ from the \texttt{gate\_proj} 
matrix and a row $W^{\text{up}}_i$ from the \texttt{up\_proj} matrix. 
Because the gating mechanism requires paired pruning 
\cite{guo_dependency-aware_2024}, every criterion combines the two rows 
into a single importance score per neuron. Neurons with the lowest scores 
are removed.

The four criteria are defined as follows:

\paragraph{PPM (Peak-to-Peak Magnitude), selected.} The sum, over both 
projections, of the peak-to-peak magnitude (maximum plus absolute minimum) 
of each weight row:
\begin{equation}
\text{PPM}_i = \Big(\max_j W^{\text{gate}}_{ij} + \big|\min_j W^{\text{gate}}_{ij}\big|\Big) 
+ \Big(\max_j W^{\text{up}}_{ij} + \big|\min_j W^{\text{up}}_{ij}\big|\Big)
\end{equation}

\paragraph{L2 (Sum of L2 Norms).} Standard norm-based magnitude pruning, 
adapted to the paired GLU setting:
\begin{equation}
\text{L2}_i = \big\|W^{\text{gate}}_i\big\|_2 + \big\|W^{\text{up}}_i\big\|_2
\end{equation}

\paragraph{VOW (Variance of Weights).} The sum of the variances of each 
weight row:
\begin{equation}
\text{VOW}_i = \mathrm{Var}\big(W^{\text{gate}}_i\big) + \mathrm{Var}\big(W^{\text{up}}_i\big)
\end{equation}

\paragraph{PON (Product of L1 Norms).} The product of the $\ell_1$ norms 
of the two weight rows:
\begin{equation}
\text{PON}_i = \big\|W^{\text{gate}}_i\big\|_1 \cdot \big\|W^{\text{up}}_i\big\|_1
\end{equation}

Among these, L2 corresponds to standard norm-based magnitude pruning, 
while VOW and PON are additional weight-only variants. We do not compare 
against activation-aware criteria such as Wanda \cite{sun_simple_2024} or 
SliceGPT \cite{ashkboos_slicegpt_2024}; that comparison is left for future 
work.

\begin{table}[H]
\caption{Catastrophic Collapse of Alternative Weight-Only Pruning Criteria (10\% Pruning, Llama-3.2-1B).}
\label{tab:method-collapse}
\centering
\resizebox{\textwidth}{!}{%
\begin{tabular}{lcccc}
\toprule
& \multicolumn{2}{c}{\textbf{WikiText (PPL)}} & \multicolumn{2}{c}{\textbf{LAMBADA (PPL)}} \\
\cmidrule(lr){2-3} \cmidrule(lr){4-5}
\textbf{Selection Criterion} & \textbf{Value} & \textbf{$\Delta$ vs. Base} & \textbf{Value} & \textbf{$\Delta$ vs. Base} \\
\midrule
Baseline (0\%)                    & 11.96   & --        & 5.42     & --         \\
PPM (Peak-to-Peak Magnitude)      & 18.37   & +54\%     & 17.89    & +230\%     \\
VOW (Variance of Weights)         & 58.03   & +385\%    & 407.21   & +7{,}400\% \\
L2 (Sum of L2 Norms)              & n/a$^*$ & n/a$^*$   & 551.10   & +10{,}100\%\\
PON (Product of L1 Norms)         & n/a$^*$ & n/a$^*$   & 1522.95  & +28{,}000\%\\
\bottomrule
\end{tabular}%
}
\\[0.5em]
\small $^*$n/a: WikiText perplexity diverged (numerical overflow).
\end{table}

Source: Preliminary experiment documented in $\text{00\_Neuron\_Selection\_Method\_Comparison.ipynb}$ (project repository). Evaluations conducted on full benchmarks.

As the data demonstrates, VOW, PON and L2 disrupt model coherence even at minimal pruning levels, increasing LAMBADA perplexity by two and three orders of magnitude. This catastrophic collapse precluded their use for gradual analysis of expansion ratios, which requires maintaining basic model functionality across multiple pruning stages ($10\%$, $20\%$, $30\%$, etc.).

The empirical superiority of PPM in preserving model stability under pruning justified its selection as the only viable criterion for this study (see Table 9). Consequently, all findings reported in this work---including the Capability Dichotomy (Section \ref{sec:dicothomy}), the inverse MMLU–TruthfulQA-MC2 correlation (Section \ref{sec:paradox}), and the energy efficiency analysis (Section \ref{sec:efficiency})---are specifically attributable to the nature of PPM-guided pruning. These findings should not be generalized to width pruning as a general technique nor to other neuron selection criteria.

\end{document}

%% file: math_commands.tex
\usepackage{amsmath,amsfonts,bm}

\def\eqref#1{equation~\ref{#1}}

\def\1{\bm{1}}

\DeclareMathAlphabet{\mathsfit}{\encodingdefault}{\sfdefault}{m}{sl}
\SetMathAlphabet{\mathsfit}{bold}{\encodingdefault}{\sfdefault}{bx}{n}

